\documentclass[journal]{IEEEtran}
\usepackage{amsmath,amsfonts}
\usepackage{algorithmic}
\usepackage{algorithm}
\usepackage{array}
\usepackage[caption=false,font=normalsize,labelfont=sf,textfont=sf]{subfig}
\usepackage{textcomp}
\usepackage{stfloats}
\usepackage{url}
\usepackage{verbatim}
\usepackage{graphicx}
\usepackage{cite}
\usepackage{lipsum}
\usepackage{hyperref}
\usepackage{tikz}
\usepackage{amssymb}
\usepackage{tablefootnote}
\usepackage{longtable}
\usepackage{listings}
\usepackage{scalerel}
\usepackage{graphicx}

\usepackage{hyperref}
\hypersetup{
    colorlinks=true, 
    linkcolor=blue,   
    citecolor=blue,   
    filecolor=blue,   %
    urlcolor=blue 
}

\newcommand{\orcidicon}{
    \scalerel*{
        \includegraphics{orcid.pdf}
    }{A}
}

\newcommand\orcid[1]{\href{https://orcid.org/#1}{\orcidicon}}

\begin{document}

\title{Spannotation: Enhancing Semantic Segmentation for Autonomous Navigation with Efficient Image Annotation}

\author{Samuel O. Folorunsho \orcid{0000-0001-6386-9190}, \IEEEmembership{Member, IEEE},
        and William R. Norris \orcid{0000-0002-4940-4458}, \IEEEmembership{Member, IEEE}}

\markboth{IEEE Transactions on Image Processing,~Vol.~X, No.~Y, February, 2024}%
{Folorunsho \MakeLowercase{\textit{et al.}}: Spannotation: Enhancing Semantic Segmentation for Autonomous Navigation with Efficient Image Annotation}

\maketitle

\begin{abstract}
Spannotation is an open source user-friendly tool developed for image annotation for semantic segmentation specifically in autonomous navigation tasks. This study provides an evaluation of Spannotation, demonstrating its effectiveness in generating accurate segmentation masks for various environments like agricultural crop rows, off-road terrains and urban roads. Unlike other popular annotation tools that requires about 40 seconds to annotate an image for semantic segmentation in a typical navigation task, Spannotation achieves similar result in about 6.03 seconds. The tools utility was validated through the utilization of its generated masks to train a U-Net model which achieved a validation accuracy of 98.27\% and mean Intersection Over Union (mIOU) of 96.66\%. The accessibility, simple annotation process and no-cost features have all contributed to the adoption of Spannotation evident from its download count of 2098 (as of February 25, 2024) since its launch. Future enhancements of Spannotation aim to broaden its application to complex navigation scenarios and incorporate additional automation functionalities. Given its increasing popularity and promising potential, Spannotation stands as a valuable resource in autonomous navigation and semantic segmentation. For detailed information and access to Spannotation, readers are encouraged to visit the project's GitHub repository at \url{https://github.com/sof-danny/spannotation} 
\end{abstract}


\section{Introduction}
\IEEEPARstart{S}{emantic} segmentation in vision-based navigation has emerged as a pivotal technology in the evolution of autonomous systems, particularly in the context of autonomous vehicle navigation, robotics, and environmental monitoring. An important first step of this technology is data gathering and data annotation. The data annotation step has often been characterized by its labor-intensive nature, time consumption, and requirement for specialized skills \cite{hassanzadeh2011machine, russo2021value, schreiner2006using, tang2012automatic, rasmussen2022challenge, hua2021semantic,vuadineanu2022analysis, bearman2016s}. The precision and accuracy in data annotation significantly contribute to the effectiveness of classification algorithms, as highlighted by \cite{fusic2021, alhazmi2021effects, ma2022effect, rasmussen2022challenge}. Having quality annotated data is thus an important first step for any machine learning model especially for semantic segmentation. 

However, current annotation tools, while diverse, come with inherent limitations. Many such tools are general-purpose, lacking the specificity required for navigation tasks, thereby complicating the annotation process \cite{siqveland, dasiopoulou2011survey, yu2018methods, finlayson2017overview}. Others demand complex usage skills or require payment and consistent internet connectivity, which restrict their practicality \cite{gibran2021}.


Recently, there have been application of the use of neural network based approach for enhanced accuracy and efficiency in data annotation \cite{shin2021all, cheema2019semantic, yu2021superpixel, zhu2020endface} for semantic segmentation and other purposes. Despite their advantages, these methods usually require some level of technical know-how, relatively high computational costs for developers and possibly cost for usage. Spannotation addresses these limitations by offering a simpler, yet efficient alternative for scenarios where intricate machine learning solutions may be impractical.

The role of data annotation in semantic segmentation extends beyond mere labelling; it is the foundation for model training, especially in vision-based navigation systems. The importance of semantic segmentation in navigation is underscored by its ability to accurately differentiate between drivable and non-drivable regions. Several research utilized semantic segmentation for different navigation tasks showing its importance and the need for quality data annotation in this area of research. Readers are refereed to works that utilized semantic segmentation for autonomous navigation in on-road \cite{miyamoto2019vision, rashed2019motion, ishida2019intersection}, off-road \cite{miyamoto2019vision, maturana2018real, ishida2019intersection}, indoor \cite{teso2020semantic, kim2018indoor, dang2023multi, Shen2023}, farm-terrains \cite{aghi2021deep, folorunsho2024semantic}.   The limitations of current annotation methods and tools act as a bottleneck in the development and implementation of effective autonomous navigation systems.

In response to these challenges, 'Spannotation' \cite{spannotation_pypi} is introduced as an innovative algorithm designed to transform the annotation process for semantic segmentation in navigation tasks. Tailored specifically for binary segmentation, Spannotation excels at distinguishing between drivable and non-drivable regions across diverse environments, including off-road terrains, crop rows for small robots, and standard on-road scenarios. Its key advantages include being fast, easy to use, free, and requiring no technical know-how, making it an ideal tool for a wide range of users.

This paper is structured to offer a comprehensive insight into Spannotation and its applications in semantic segmentation for vision-based navigation. Section [\ref{section2}] spells out the technical aspects and innovative approaches employed in Spannotation. In the Results section [\ref{results}], the algorithm's efficacy is shown through various test cases and comparisons with existing tools. The ensuing Discussion delves into the implications of these results and the potential impact of Spannotation in the field. Lastly, the paper concludes [\ref{section_conclude}] with a summary of findings and perspectives on future enhancements and applications of this tool.

\section{Methodology}
\label{section2}
\subsection{Spannotation Algorithm}
\begin{figure*}[!t]
\centering
\includegraphics[width=\textwidth]{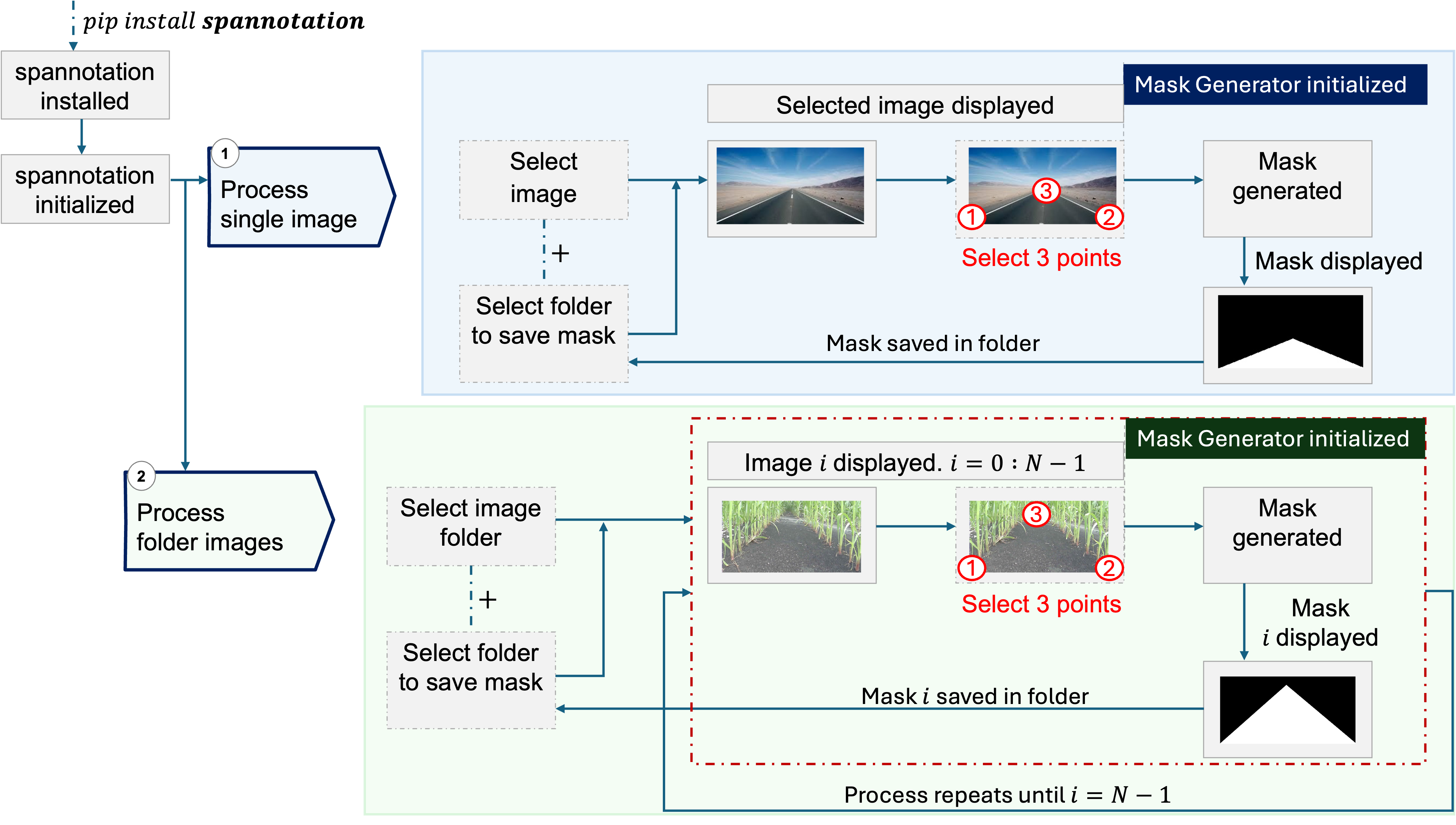}
\caption{General overview of the Spannotation process. Spannotation can be used for either single image (option 1) or multiple images in a folder (option 2) where N is the total number of images in the folder. There is no order to which the three (3) points can be selected on the image.}
\label{fig:spannotation_overview}
\end{figure*}

The Spannotation algorithm operates through a user-friendly interface, allowing users to annotate images for semantic segmentation by manually selecting three (3) points of interest. The workflow of Spannotation is shown in Fig. \ref{fig:spannotation_overview}.

\begin{algorithm}[H]
\caption{Spannotation Algorithm}\label{alg:spannotation}
\begin{algorithmic}
\STATE 
\STATE \textbf{Input:} Image set $I$, Output path $O$
\STATE \textbf{Output:} Segmentation masks for each image in $I$
\STATE
\STATE \textsc{Spannotation}$(I, O)$
\STATE \hspace{0.5cm}\textbf{for} each image $i \in I$ \textbf{do}
\STATE \hspace{1.0cm}Initialize empty list $P$ for points
\STATE \hspace{1.0cm}$maskGenerated \gets \text{False}$
\STATE \hspace{1.0cm}Display image $i$ for annotation
\STATE \hspace{1.0cm}\textbf{while} not $maskGenerated$ \textbf{do}
\STATE \hspace{1.5cm}Capture user input points $(x, y)$
\STATE \hspace{1.5cm}Add $(x, y)$ to $P$
\STATE \hspace{1.5cm}\textbf{if} $\text{length}(P) = 3$ \textbf{then}
\STATE \hspace{2.0cm} $mask \gets \text{GenerateMask}(P, i)$
\STATE \hspace{2.0cm}Save $mask$ to path in $O$
\STATE \hspace{2.0cm}$maskGenerated \gets \text{True}$
\STATE \hspace{1.5cm}\textbf{end if}
\STATE \hspace{1.0cm}\textbf{end while}
\STATE \hspace{0.5cm}\textbf{end for}
\STATE
\STATE \textsc{GenerateMask}$(P, i)$
\STATE \hspace{0.5cm}Create blank mask $M$ with dimensions of $i$
\STATE \hspace{0.5cm}Fill polygon defined by $P$ in $M$ with white
\STATE \hspace{0.5cm}\textbf{return} $M$
\end{algorithmic}
\end{algorithm}

The process begins with the initialization of the MaskGenerator class, which will handle the creation of binary masks. It includes an event handler that responds to user clicks within the image window. The user's first three clicks define the vertices of a triangular region of interest. Upon the third click, a binary mask is generated where the selected triangular region is filled with white on a black background, indicating the drivable area. This mask is then displayed to the user, saved to a specified directory, and the process can be repeated for additional images or terminated by pressing 'q' on the keyboard. The class also contains methods for processing a single image or all images within a folder, streamlining the annotation task for large datasets.

\subsection{Theoretical Mathematics Behind Spannotation}

Spannotation employs fundamental concepts from computational geometry to generate binary masks for semantic segmentation. The process involves constructing geometric shapes based on user-defined points and then rasterizing these shapes onto a binary image where the foreground (drivable area) and background (non-drivable area) are clearly defined.

\subsubsection{Geometric Primitives}

The core geometric primitive used in Spannotation is the triangle, the simplest polygon, which is defined by three non-collinear points. Given points \( P_1, P_2, \) and \( P_3 \), a triangle can be mathematically represented as the set of all points \( P \) that satisfy the following condition:

\begin{equation}
P = \alpha P_1 + \beta P_2 + \gamma P_3
\end{equation}

where \( \alpha, \beta, \) and \( \gamma \) are barycentric coordinates such that \( \alpha, \beta, \gamma \geq 0 \) and \( \alpha + \beta + \gamma = 1 \) \cite{farin2002,pottmann2002geometries,cheutet2007constraint}.

\subsubsection{Rasterization}

Rasterization is the process of converting the geometric description of a shape into a binary image \cite{pineda1988parallel, gharachorloo1989characterization}. It involves filling the pixels within the shape's boundaries. For a triangle, this process can be described using the following function:

\begin{equation}
f(x, y) = 
\begin{cases} 
1 & \text{if } (x, y) \text{ is in the $\triangle$ defined by } P_1, P_2, P_3 \\
0 & \text{otherwise}
\end{cases}
\end{equation}

The triangle rasterization algorithm implemented in Spannotation uses scanline conversion \cite{watt1993}, which is an efficient way to fill the triangle by intersecting it with horizontal lines (scan lines) and filling the segments of the scan lines that lie inside the triangle .

\subsubsection{Mask Generation}

The mask generation in Spannotation is thus a practical application of these theoretical principles. The user-defined points act as vertices for triangles, and the rasterization algorithm fills these triangles to create the mask. The binary mask \( M \) for an image \( I \) with dimensions \( w \times h \) is represented as:

\begin{equation}
M_{i,j} = f(i, j) \quad \text{for } i = 1, \ldots, w \text{ and } j = 1, \ldots, h
\end{equation}

where \( M_{i,j} \) is the value of the mask at pixel location \( (i, j) \), which is 1 if the pixel is within the triangle and 0 otherwise~\cite{szeliski2010}.

\subsection{Dependencies and Their Functions}

Spannotation relies on several external libraries, each serving a different purpose in the process of image annotation for semantic segmentation. The following subsections detail the roles of these dependencies.

\subsubsection{OpenCV (cv2)}

OpenCV \cite{opencv_library}, abbreviated as cv2 in Python, is an open-source computer vision and machine learning software library. Spannotation utilizes OpenCV for a multitude of tasks:

\begin{itemize}
    \item \textbf{Image Handling:} Reading, displaying, and writing images.
    \item \textbf{Graphical User Interface (GUI):} Creating windows to display images and capture user inputs through mouse clicks.
    \item \textbf{Drawing Functions:} Marking points and drawing shapes (triangles) on images based on user selection.
    \item \textbf{Mask Operations:} Generating binary masks with the filled polygon areas defined by user-selected points.
\end{itemize}

OpenCV's comprehensive set of functions allows Spannotation to handle these image processing tasks with relative ease~\cite{bradski2000opencv}.

\subsubsection{NumPy}

NumPy \cite{harris2020array}, or Numerical Python, is a foundational package for scientific computing in Python. Spannotation employs NumPy primarily for:

\begin{itemize}
    \item \textbf{Array Manipulation:} Representing images and masks as multi-dimensional arrays for efficient processing.
    \item \textbf{Mathematical Operations:} Performing calculations needed for mask generation.
\end{itemize}

NumPy's array object is much more efficient than native Python lists, especially for operations on large datasets, making it ideal for image data manipulation \cite{oliphant2006guide}.

\subsubsection{OS}

The os module \cite{python_os} in Python provides a way of using operating system-dependent functionality. Spannotation uses the os module to:

\begin{itemize}
    \item \textbf{File Path Handling:} Managing file paths for reading and saving images and masks.
    \item \textbf{Directory Management:} Creating directories for saving the generated masks.
\end{itemize}

The os module ensures that Spannotation's file operations are portable across different operating systems \cite{drake2011python}.

\subsubsection{Glob}

The glob module \cite{python_glob} finds all the pathnames matching a specified pattern according to the rules used by the Unix shell. In Spannotation, glob is used to:

\begin{itemize}
    \item \textbf{File Retrieval:} Accessing all image files within a specified directory, enabling batch processing of images for mask generation.
\end{itemize}

Glob simplifies the task of file retrieval, allowing Spannotation to process multiple images in a user-friendly manner~\cite{drake2011python}.

\subsection{Sample Implementation of Spannotation}

Spannotation simplifies the process of generating masks for semantic segmentation tasks and can be easily installed and used both in a Python code editor and through the command line or terminal.

\subsubsection{Installation}
\label{install}
To install Spannotation, use pip with the following command:

\begin{verbatim}
pip install spannotation
\end{verbatim}

To ensure the latest version installed, run:

\begin{verbatim}
pip install --upgrade spannotation
\end{verbatim}

Alternatively, to install a specific version of Spannotation, use:

\begin{verbatim}
pip install spannotation==0.1.11 
\end{verbatim}

where 0.1.11 is the version number. 

\subsubsection{Usage in a Python Code Editor}

After installation, Spannotation can be utilized within a typical code editor environment like Jupyter Notebook or Visual Studio Code as follows:

\paragraph{Step 1: Install Spannotation.}

Ensure spannotation is installed as described in [\ref{install}]. 

\paragraph{Step 2: Import the Package.}

Import the MaskGenerator class from Spannotation at the beginning of the script:

\begin{verbatim}
from Spannotation import MaskGenerator
\end{verbatim}

\paragraph{Step 3: Initialize the Generator.}

Instantiate the MaskGenerator:

\begin{verbatim}
generator = MaskGenerator()
\end{verbatim}

\paragraph{Step 4: Process Images.}

Use the generator to process images. For a single image:

\lstset{
  basicstyle=\ttfamily\footnotesize,
  breaklines=true,
  frame=none,
  captionpos=b
}

\begin{lstlisting}[language=Python]
generator.process_image(path/to/image, 'path/to/save/masks')
\end{lstlisting}

For processing all images in a folder:

\begin{lstlisting}[language=Python]
generator.process_folder('path/to/folder', 'path/to/save/masks')
\end{lstlisting}

Replace the paths with the actual locations on the system where images are stored and where the generated masks should be stored.

\subsubsection{Usage in Command Line / Terminal}

Spannotation can also be operated through the command line or terminal after installation.

\paragraph{Process a Single Image.}

To generate a mask for a single image:

\begin{enumerate}
    \item Run the command:
    \begin{verbatim}
    python3 -m Spannotation.cli
    \end{verbatim}
    \item Choose option 1 for a single image.
    \item Enter the full path to the image.
    \item Enter the full path where the mask should be saved.
\end{enumerate}

\paragraph{Process Multiple Images in a Folder.}

To process multiple images:

\begin{enumerate}
    \item Run the command:
    \begin{verbatim}
    python3 -m Spannotation.cli
    \end{verbatim}
    \item Choose option 2 for a folder.
    \item Enter the full path to the folder containing images.
    \item Enter the full path where the masks should be saved.
\end{enumerate}

The command line interface provides an easy-to-follow menu that guides through the mask generation process for either single images or batches of images within a directory without the need to install code editors.

\subsection{Method for Testing the Algorithm}

To evaluate the effectiveness of the Spannotation algorithm, tests were conducted across three distinct navigation environments: agricultural row crops, off-road trails, and on-road scenarios. These environments present unique challenges for semantic segmentation due to varying textures, lighting conditions, and object arrangements. For each environment, the following steps were followed:

\begin{enumerate}
    \item A set of representative images was selected for each environment.
    \item Spannotation was utilized to annotate the images, designating drivable and non-drivable regions by selecting three (3) key points to form polygons that define these areas.
    \item The generated masks were saved and reviewed for accuracy and completeness.
\end{enumerate}

The resulting masks from these tests are shown in the section [\ref{result_2}].

\subsection{Training with U-Net Architecture}
Semantic segmentation is important in vision-based navigation. The accuracy of such systems is heavily reliant on the quality of the segmentation masks used during the training phase of the segmentation model. These masks inform the model about the precise boundaries between different classes within an image, which, in the context of autonomous navigation, typically translates into distinguishing between drivable and non-drivable areas. 

\par To ascertain the efficacy of Spannotation in real-world applications, it is important to verify that the masks it generates can indeed serve as effective training data. The premise is straightforward: if a segmentation model trained on Spannotation-generated masks can accurately segment new, unseen images, then one can affirm the effectiveness of Spannotation.

\subsubsection{Dataset Description}
The dataset employed in this study is comprised of original images of corn rows, provided by \cite{sivakumar2021learned}, containing a total of 1030 images. This dataset was captured by the EarthSense robot \cite{earthsense_website}. 

\subsubsection{U-Net Architecture}
For the semantic segmentation tasks, the U-Net architecture was employed. U-Net is a convolutional neural network originally designed for biomedical image segmentation \cite{ronneberger2015u}. U-Net is particularly suited for these needs due to its efficacy with smaller sample sizes and its precision in segmentation.

The U-Net model features a U-shaped architecture composed of a contracting path to capture context and an expansive path that enables precise localization. The contracting path consists of repeated application of two 3x3 convolutions, each followed by a ReLU and a 2x2 max pooling operation. The expansive path includes upsampling of the feature map followed by a 2x2 convolution that halves the number of feature channels, and a concatenation with the correspondingly cropped feature map from the contracting path. The architecture culminates with a 1x1 convolution that maps the features to the desired number of classes.

\begin{figure}[!t]
\centering
\includegraphics[width=3.5in]{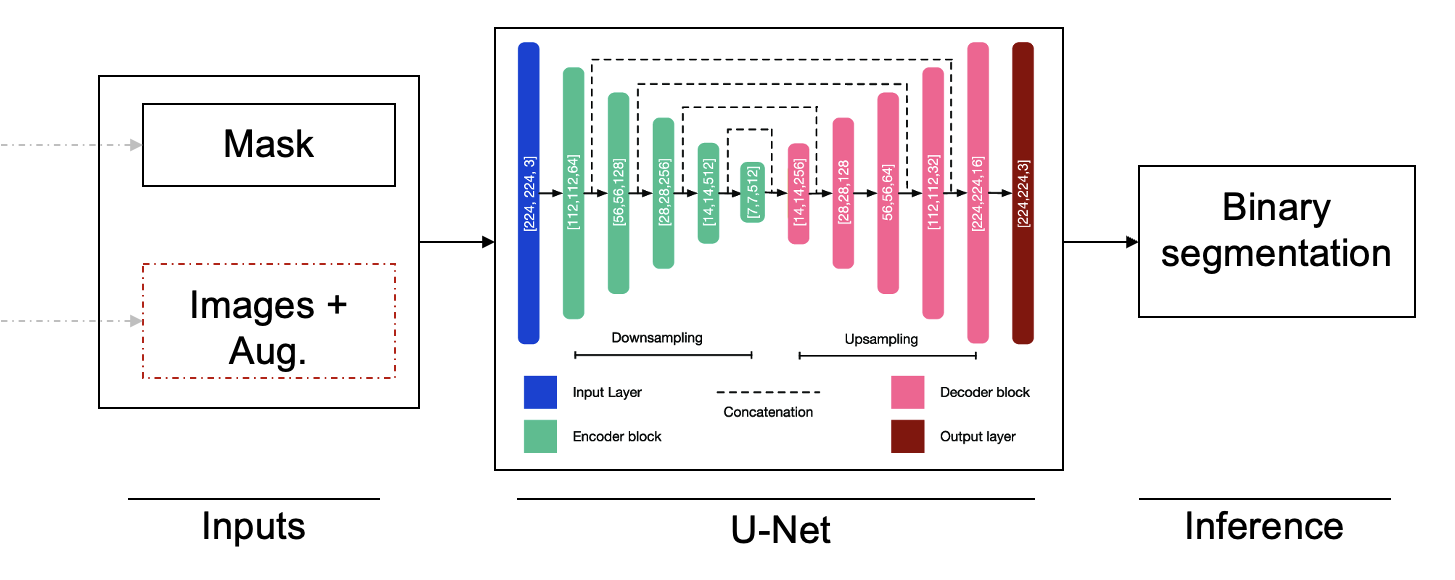}
\caption{Schematic representation of the U-Net architecture.}
\label{unet}
\end{figure}

Fig. \ref{unet} shows the U-Net architecture, demonstrating the flow from the input image and mask to the resultant binary mask segmentation.

\subsubsection{Training Process}
The training of the U-Net model involved augmentation of the dataset to ensure robustness against various environmental conditions. Techniques such as horizontal flipping, rotation, color jittering, cropping, exposure adjustment, and blurring were applied to augment the images.

Training was executed on the UIUC NCSA HAL cluster~\cite{kindratenko2020hal}, utilizing its GPU capabilities for enhanced performance. The model was trained with the following parameters:

\begin{itemize}
    \item Number of epochs: 33
    \item Optimizer: Adam \cite{kingma2014adam}
    \item Learning rate: 0.001
    \item Loss function: Binary Cross-Entropy
\end{itemize}

The dataset was partitioned into an 85\% (876 images) training set and a 15\% (154 images) validation set.

\subsubsection{Accuracy Metrics}
The model's performance was evaluated using validation loss, validation accuracy, and mean Intersection over Union (mIOU) as metrics. These metrics provided insights into the model's prediction accuracy, its ability to generalize, and the quality of the segmentation in comparison to the ground truth annotations.

\begin{figure}[!t]
\centering
\includegraphics[width=3.5in]{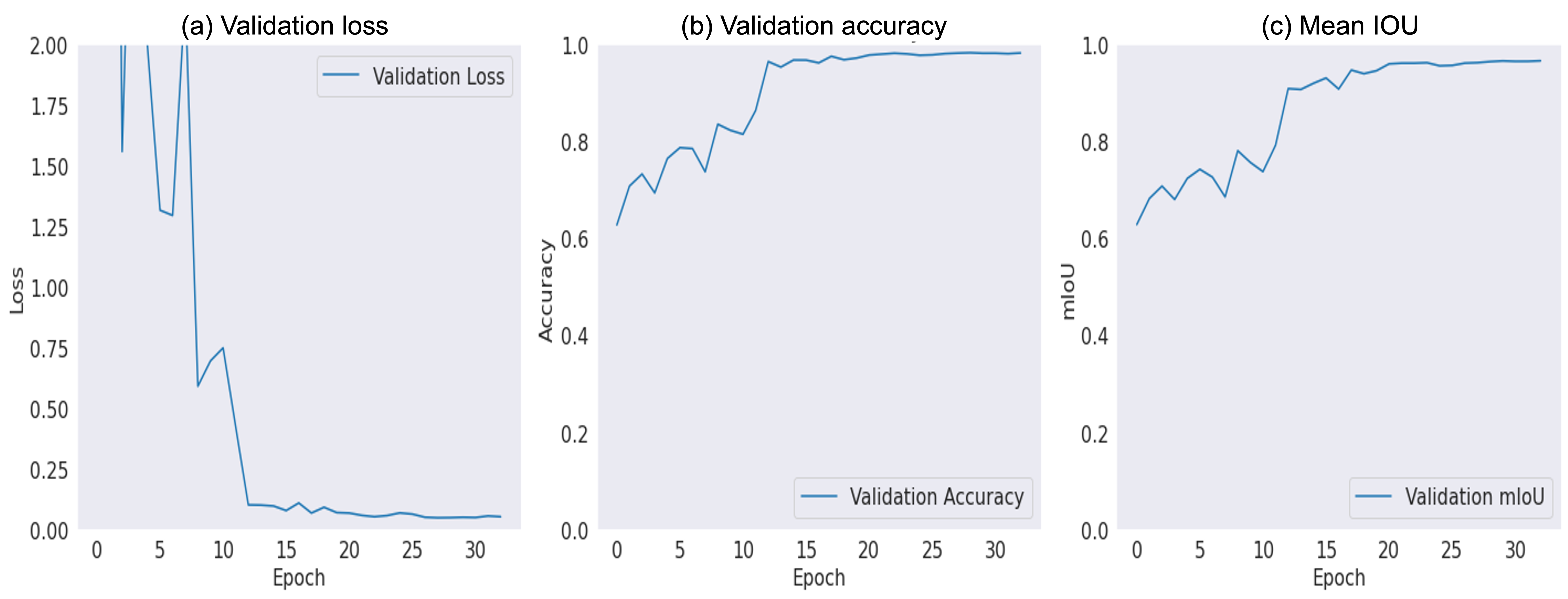}
\caption{U-Net training results over 33 epochs: (a) Validation loss, (b) Validation accuracy, and (c) Mean Intersection over Union (mIOU).}
\label{fig:unet_training_results}
\end{figure}

The Fig. \ref{fig:unet_training_results} shows the training results and Fig. \ref{segmentation} shows the segmentation results.

\section{Results and Discussion}
\label{results}
In this section, the results of all the tests described in the methodology are presented. 

\subsection{Spannotation mask generation}
\label{result_2}
\begin{figure}[!t]
\centering
\includegraphics[width=3.5in]{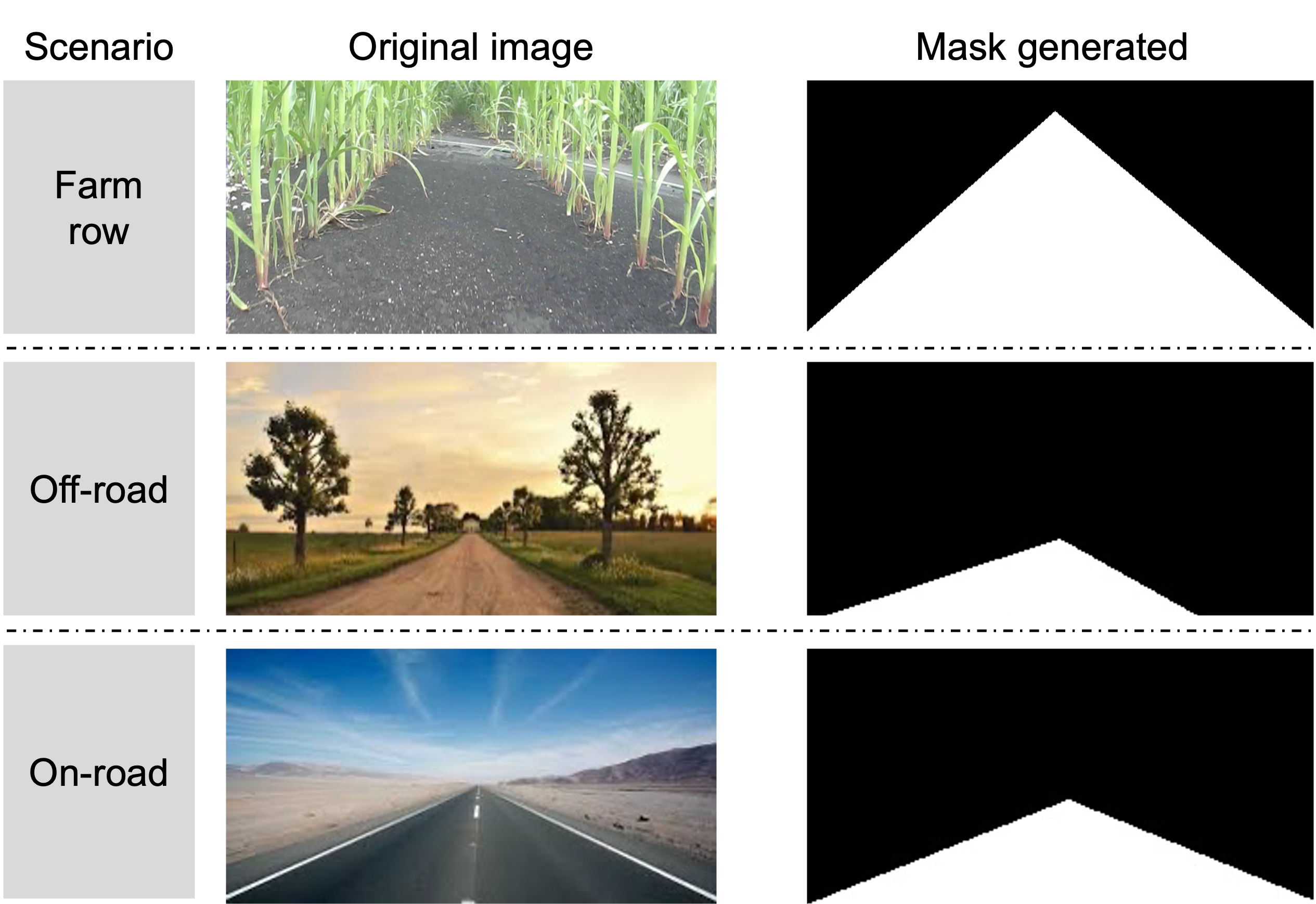}
\caption{Original images and corresponding masks generated by Spannotation for different scenarios: agricultural crop rows, off-road, and on-road.}
\label{mask_1}
\end{figure}

The application of Spannotation across different environments is shown in Fig.\ref{mask_1}. The algorithm was tested on images from agricultural crop rows, off-road trails, and on-road scenarios.

The consistency of the masks with the target regions across diverse environments shows the robust annotation capabilities of Spannotation. 

\subsection{U-Net Training Results}
\label{result_3}
The training of the U-Net model with Spannotation-generated masks yielded results in terms of model performance metrics. After 33 epochs of training, the model achieved a validation loss of 0.0506, a validation accuracy of $98.27\%$, and a mean Intersection over Union (mIOU) of $96.66\%$ as shown in Fig.\ref{segmentation}.

The results from the U-Net training shows the quality of the masks generated by Spannotation, which was important in guiding the model towards the accuracy and precision shown in the figure.

\subsection{U-Net Segmentation Results}

Following the successful training of the U-Net model, a segmentation test was conducted on test images to evaluate the model's practical performance. The results shows a strong similarity between the U-Net-generated segmentation masks and the original masks created by Spannotation (used for training).

\subsubsection{Comparison of Segmentation Masks}

\begin{figure}[!t]
\centering
\includegraphics[width=3.5in]{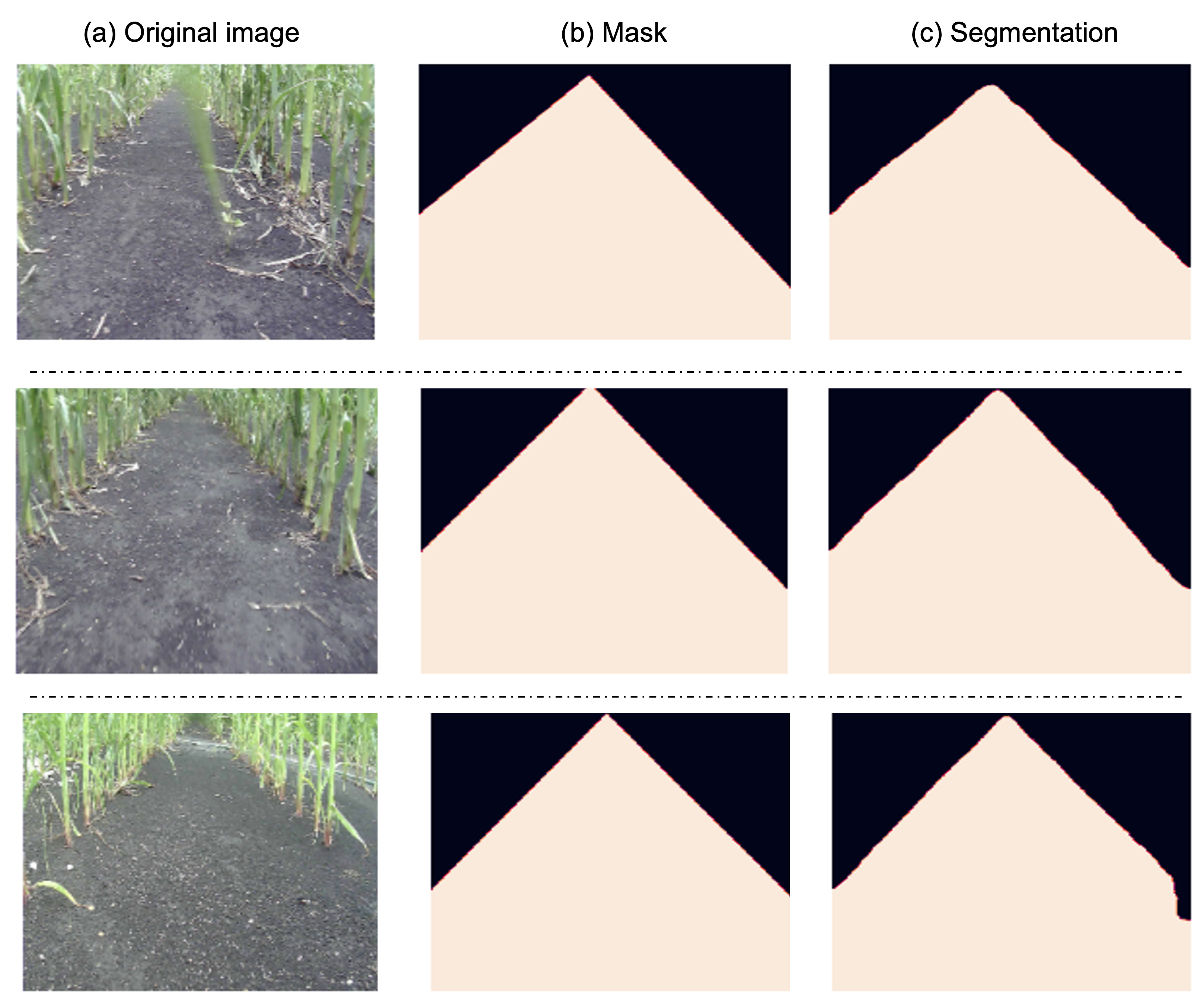}
\caption{Comparison of segmentation results: (a) Original images (b) spannotation masks, and (c) U-Net generated segmentation.}
\label{segmentation}
\end{figure}

The comparison of the segmentation results from the U-Net model with the original Spannotation masks shows a high degree of similarity. Fig. \ref{segmentation} presents a side-by-side comparisons of the original images, the Spannotation masks, and the U-Net generated masks. Both masks closely traced the drivable and non-drivable regions in the original images.

This visual comparison clearly illustrates the U-Net model's capability to accurately segment images based on the training it received from the Spannotation-generated masks.

\subsection{Comparison of Spannotation with Other Annotation Tools}

A comparative analysis was conducted to evaluate the efficacy of Spannotation against other prevalent image annotation tools. This comparison is essential to understand how Spannotation stands in terms of setup time, annotation speed, cost, technical requirements, and its suitability for navigation tasks.

\subsubsection{Criteria for Comparison}

The comparison was based on the following criteria:

\begin{enumerate}
    \item \textbf{Setup Time:} The total time required to get the tool ready for use, including installation and initial configuration.
    \item \textbf{Time to Annotate One Image:} The average time taken to annotate a single image.
    \item \textbf{Cost:} The monetary cost of using the tool.
    \item \textbf{Additional Technicalities:} Any extra technical steps or knowledge required for using the tool.
    \item \textbf{Usability for Navigation Task:} Suitability of the tool for annotating images specifically for navigation tasks.
\end{enumerate}

\subsubsection{Comparative Analysis}
Table \ref{tab:annotation_tools_comparison} presents a comparison of Spannotation with other popular annotation tools \cite{label_top10_2023} using the 1030 row crop data made available by \cite{sivakumar2021learned}.

\begin{table*}[htbp]
\caption{Comparison of Spannotation with other annotation tools}
\begin{center}
\renewcommand{\arraystretch}{1.2} 
\begin{tabular}{|l|c|c|c|c|}
\hline
\textbf{Criteria} & \textbf{Spannotation} & \textbf{Label Box \textsuperscript{*}\cite{labelbox_pricing}} & \textbf{Labellerr \textsuperscript{*}} \cite{labellerr_pricing} & \textbf{VGG Annotator } \cite{vgg_via_demo}\\
\hline
Set Up Time (seconds) & \textbf{148.2}\textsuperscript{1} & 384 \textsuperscript{2}& 609 \textsuperscript{3}& 155.4 \textsuperscript{4}\\
\hline
Annotate one Image (seconds) & \textbf{6.03} & 55 & 40.6 & 39\\
\hline
Cost & Free & Limited free version \textsuperscript{5} & Limited free version \textsuperscript{6} & Free\\
\hline
Additional Technicalities & No & Yes & Yes & No\\
\hline
Usability for Navigation Task & Yes & Yes & Yes & Yes\\
\hline
\multicolumn{5}{l}{\footnotesize \textsuperscript{1} \textit{navigating the GitHub page, installation, copying and pasting paths for images/folders, copying and pasting the other steps from the  GitHub page.}} \\
\multicolumn{5}{l}{\footnotesize \textsuperscript{2,3} \textit{navigating the website, account set-up, loading in the data (this depends on number of images), understanding the interface}} \\
\multicolumn{5}{l}{\footnotesize \textsuperscript{4} \textit{navigating the website,  loading in the data, items, understanding the interface}} \\
\multicolumn{5}{l}{\footnotesize \textsuperscript{5} \textit{\$8.33 per month for a typical 5000 unit of data operations. Free limited to 500 LabelBox Units(LBUs) per month \cite{labelbox_pricing}.}} \\
\multicolumn{5}{l}{\footnotesize \textsuperscript{6} \textit{\$ 299 USD per month for pro. Free version for students and researchers. Free limited to 100 Data Credit per month \cite{labellerr_pricing}.}} \\
\multicolumn{5}{l}{\footnotesize \textsuperscript{*} \textit{Have other very useful functionalities beyond data annotation including model training and AI services.}} \\
\end{tabular}
\label{tab:annotation_tools_comparison}
\end{center}
\end{table*}

This comparative analysis highlights the strengths and weaknesses of each tool, with Spannotation demonstrating significant advantages in certain aspects, particularly in ease of setup and annotation speed, specificity for navigation tasks making it a highly suitable choice for navigation-related segmentation tasks.

\subsection{Acceptance of Spannotation: Download Metrics}

An important indicator of the acceptance and utility of Spannotation in the broader community is the number of times it has been downloaded and potentially used. To assess this, data from Pepy \cite{spannotation_pepy}, a website that tracks the download statistics of Python packages distributed via PyPI (the Python Package Index) was utilized.

\subsubsection{Data Collection Method}

The website `pepy.tech` aggregates download data from PyPI's official logs, providing insights into the usage trends of Python packages. The download count reflects the number of times Spannotation has been installed or updated using the `pip` command. This metric serves as a proxy for the tool's popularity and acceptance among developers and researchers in the field of computer vision and autonomous navigation.

\subsubsection{Interpretation of Download Data}

As of the latest data, Spannotation has been downloaded a total of 2,152 times across its different versions (as at Feb. 26, 2024). This number signifies not just the initial interest but also the continued usage of Spannotation, as it includes both new installations and updates by users. A high download count typically indicates that a tool is being actively used and is deemed valuable by its user base.

\begin{figure}[!t]
\centering
\includegraphics[width=3.5in]{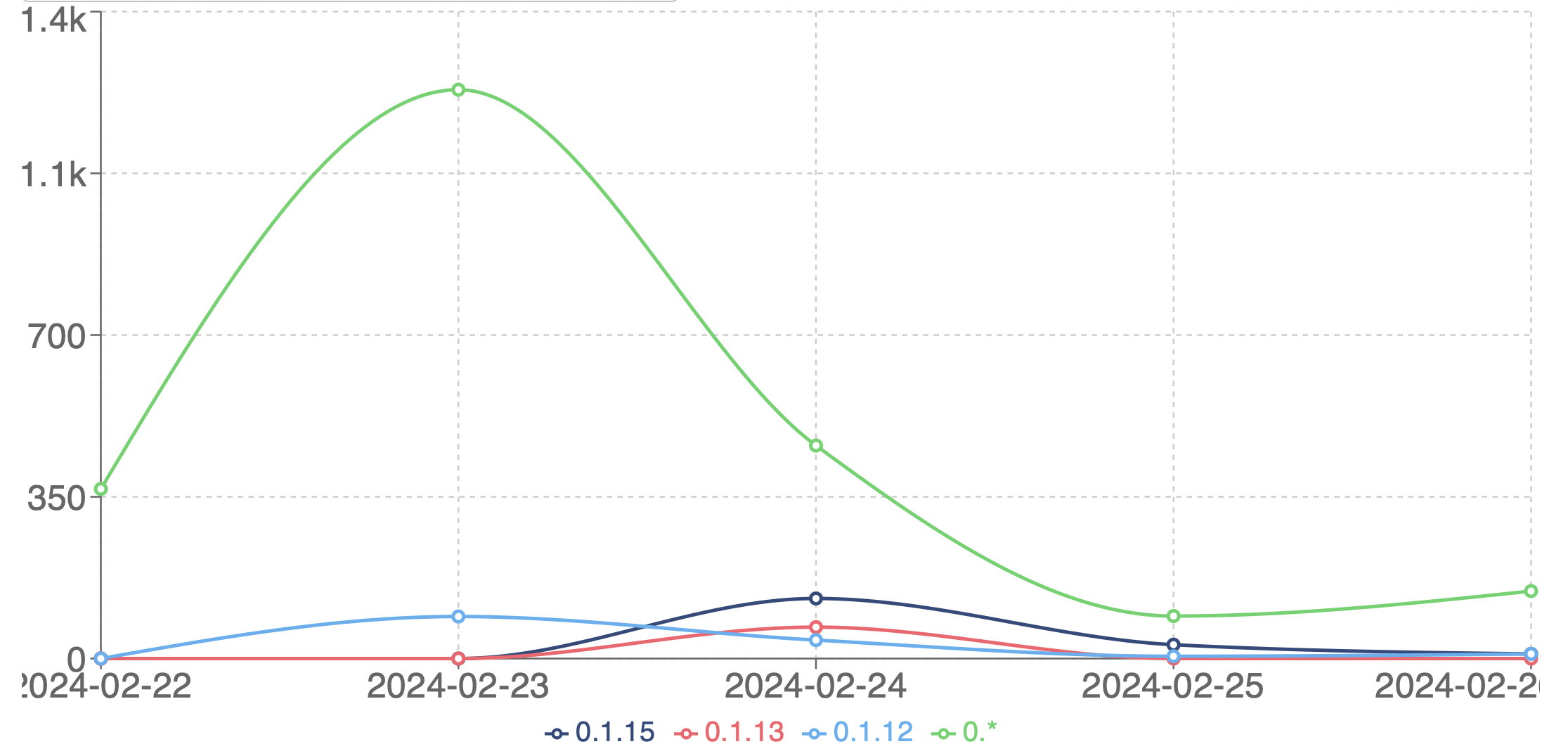}
\caption{Trend of downloads for different versions of Spannotation from its release to the present, as tracked by pepy.tech.}
\label{download_trend}
\end{figure}

Fig. \ref{download_trend} shows the trend of downloads for Spannotation since its release. The upward trajectory in the graph not only validates the tool's increasing popularity but also suggests a growing interest in tools that simplify the process of generating masks for semantic segmentation tasks.

\section{Conclusion and future direction}
\label{section_conclude}
This research introduces Spannotation, a tool for annotating images specifically tailored for semantic segmentation in navigation tasks. Through tests and comparisons Spannotation has shown its effectiveness in creating accurate masks for different settings like agricultural crop rows, off road trails and on-road situations. The successful training and validation of a U-Net model using masks generated by Spannotation further confirms the tools usefulness. Moreover the considerable number of downloads since its launch indicates a reception within the community. Spannotation emerges as a tool in semantic segmentation especially for autonomous navigation tasks. Its ability to produce masksfor training efficient segmentation models is its most notable contribution. As the tool progresses, it is anticipated to play an essential role in advancing autonomous navigation systems assisting in developing more precise and reliable models capable of handling real world complexities.

Looking forward, there are ways to enhance Spannotations capabilities. Future versions will concentrate on expanding its adaptability to address intricate navigation scenarios. This enhancement will involve customizing the tool, for usage in environments featuring curved roads, uneven terrains and dynamically changing surroundings.

Another important focus area is incorporating more automation into the labeling process. With the use of AI technology, the time and effort needed for annotation tasks can be reduced while maintaining high accuracy levels. This automated approach could play a role in expanding the use of Spannotation for larger datasets.

There are also plans to enhance the user interface. Ongoing improvements in this aspect will enhance the tools usability and accessibility making it easier for individuals without technical expertise to use it. A user friendly interface is essential for adoption and user friendliness.

Finally, prioritizing compatibility and integration with existing software tools and platforms used in autonomous navigation is crucial. This integration will enhance Spannotations versatility and ease of use positioning it as a tool in the field of autonomous navigation and semantic segmentation. These developments aim to establish Spannotation as a go-to tool, for segmentation image annotation tasks in autonomous navigation.


\bibliographystyle{IEEEtran}
\bibliography{IEEEfull, ref}

\begin{IEEEbiography}[{\includegraphics[width=1in,height=1.25in,clip,keepaspectratio]{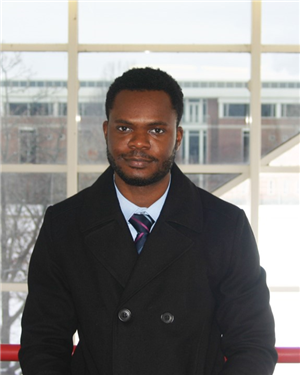}}]{SAMUEL O. FOLORUNSHO is a Graduate member of IEEE and
a PhD student conducting research at the Autonomous and
Unmanned Vehicle Systems Lab (AUVSL) and the Center for Autonomous
Construction and Manufacturing at Scale (CACMS) in the Department of
Systems Engineering at the University of Illinois, Urbana-Champaign (UIUC) under the advisorship of Prof. William R. Norris. His research is focused on control systems, computer vision and robotics - and the intersection of those for
safety-critical systems in industrial and agricultural applications.
He earned his M.S. from UIUC in 2023 and his B.S. in 2017 at
the University of Ilorin, Nigeria both in Agricultural and Biological
Systems Engineering. He has three years of working experience in management consulting.}
\end{IEEEbiography}

\vspace{11pt}

\begin{IEEEbiography}[{\includegraphics[width=1in,height=1.25in,clip,keepaspectratio]{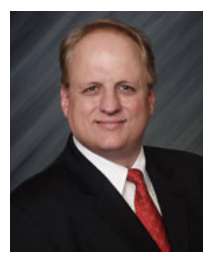}}]{WILLIAM R. NORRIS (Member, IEEE) received
the B.S., M.S., and Ph.D. degrees in systems engineering from the University of Illinois at Urbana–
Champaign, in 1996, 1997, and 2001, respectively,
and the M.B.A. degree from the Fuqua School of
Business, Duke University, in 2007. He has over
23 years of industry experience with autonomous
systems. He is currently a Clinical Associate Professor with the Industrial and Enterprise Systems
Engineering Department, University of Illinois at
Urbana–Champaign, the Director of the Autonomous and Unmanned Vehicle
System Laboratory (AUVSL), as well as the Founding Director of the Center
for Autonomous Construction and Manufacturing at Scale (CACMS).}
\end{IEEEbiography}

\vfill

\end{document}